\documentclass[sigconf]{acmart}

\copyrightyear{2024}
\acmYear{2024}
\setcopyright{acmlicensed}\acmConference[KDD '24]{Proceedings of the 30th ACM SIGKDD Conference on Knowledge Discovery and Data Mining}{August 25--29, 2024}{Barcelona, Spain}
\acmBooktitle{Proceedings of the 30th ACM SIGKDD Conference on Knowledge Discovery and Data Mining (KDD '24), August 25--29, 2024, Barcelona, Spain}
\acmDOI{10.1145/3637528.3671688}
\acmISBN{979-8-4007-0490-1/24/08}

\usepackage{booktabs} % For formal tables
\usepackage{amsmath}
\usepackage{amsthm}
\usepackage{multirow,tabularx,ragged2e}
\usepackage{adjustbox}

\newcounter{noteZZctr} 

\begin{document}

\title{Representation Learning of Geometric Trees}

\author{Zheng Zhang}
\orcid{0009-0008-9808-6020}
\affiliation{%
  \institution{Emory University}
  \city{Atlanta}
  \state{GA}
  \country{USA}
  \postcode{30324}
}
\email{zheng.zhang@emory.edu}

\author{Allen Zhang}
\affiliation{%
  \institution{Georgia Institute of Technology}
  \city{Atlanta}
  \state{GA}
  \country{USA}}
\email{azhang490@gatech.edu}

\author{Ruth Nelson}
\affiliation{%
  \institution{Yale University}
  \city{New Haven}
  \state{CT}
  \country{USA}}
\email{ruth.nelson@yale.edu}

\author{Giorgio Ascoli}
\affiliation{%
 \institution{George Mason University}
 \city{Fairfax}
 \state{VA}
 \country{USA}}
\email{ascoli@gmu.edu}

\author{Liang Zhao}
\authornote{Corresponding Author}
\affiliation{%
  \institution{Emory University}
  \city{Atlanta}
  \state{GA}
  \country{USA}
  \postcode{30324}
}
\email{liang.zhao@emory.edu}


\begin{abstract}
    Geometric trees are characterized by their tree-structured layout and spatially constrained nodes and edges, which significantly impacts their topological attributes. This inherent hierarchical structure plays a crucial role in domains such as neuron morphology and river geomorphology, but traditional graph representation methods often overlook these specific characteristics of tree structures. To address this, we introduce a new representation learning framework tailored for geometric trees. It first features a unique message passing neural network, which is both provably geometrical structure-recoverable and rotation-translation invariant. To address the data label scarcity issue, our approach also includes two innovative training targets that reflect the hierarchical ordering and geometric structure of these geometric trees. This enables fully self-supervised learning without explicit labels. We validate our method's effectiveness on eight real-world datasets, demonstrating its capability to represent geometric trees.
    
    %(((((Geometric tree graphs are characterized by their tree-structured layout and spatially constrained nodes and edges, which significantly influences their topological attributes. This naturally hierarchical structure is essential for understanding phenomena such as river geomorphology, but traditional graph representation methods often overlook the specific characteristics inherent in these tree structures. To address this, we introduce a new representation learning framework tailored for geometric tree graphs. It features a unique message passing neural network that is both provably capable of recovering the geometric structure of tree graphs and rotation-translation invariant. To enhance comprehension of the inherent tree-structured relationships, our approach also includes two innovative training targets reflecting the hierarchical and geometric nature of these tree graphs. This enables fully self-supervised learning without explicit labels. We validate our method’s effectiveness on seven real-world datasets, demonstrating its capability to represent geometric tree graphs.)
\end{abstract}




\maketitle

\section{Introduction}

\begin{figure}[t]
    \centering
    \includegraphics[width=1.\linewidth]{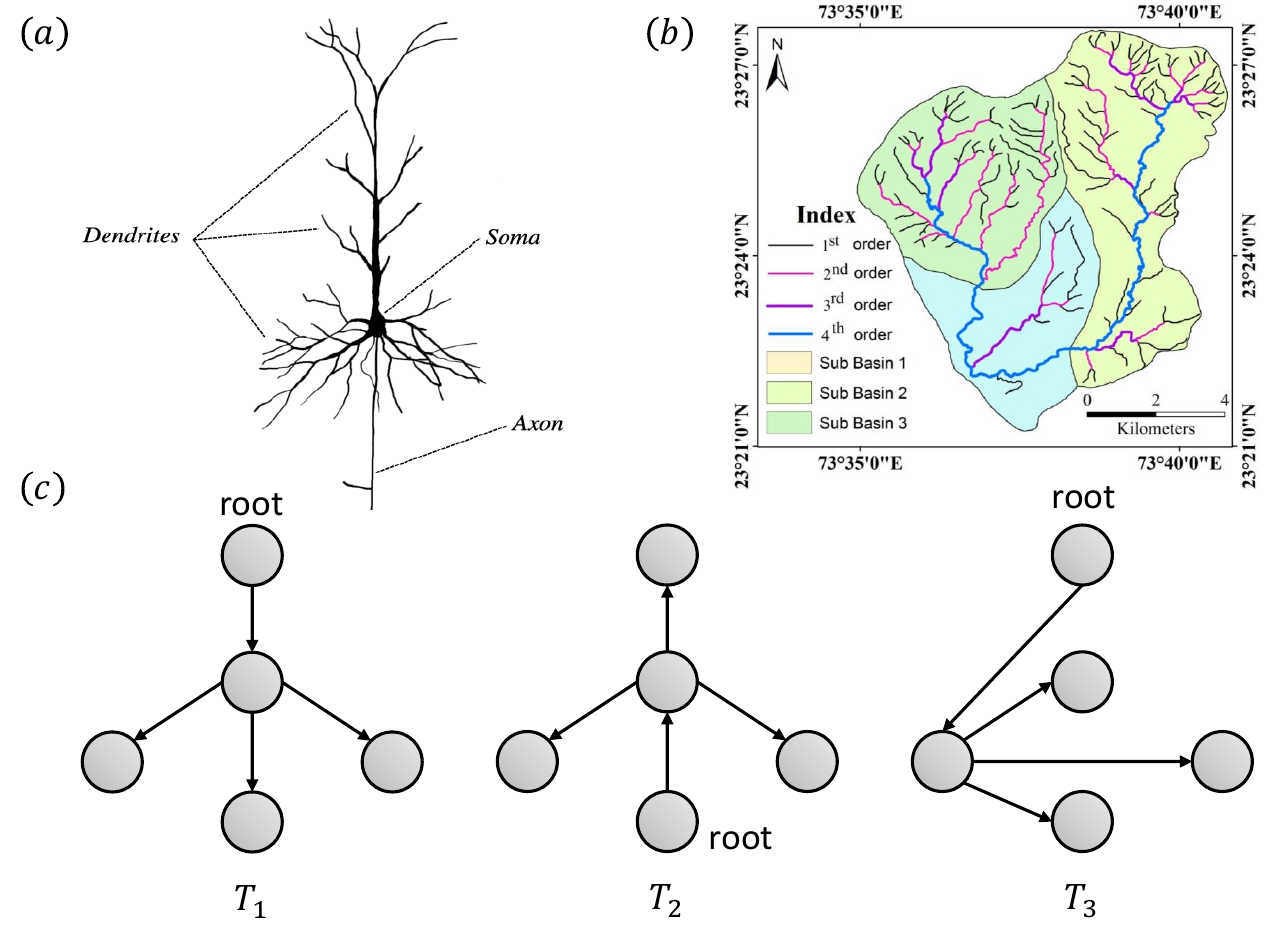}
    \vspace{-6mm}
    \caption{(a) Illustration of a neuron's geometric tree-like structure; (b) Representation of a river network exhibiting a tree structure embedded within a geometric landscape; (c) Three different geometric trees with isomorphic network connectivity and identical spatial coordinates. Distinguishing these geometric trees requires jointly considering spatial, topology, and hierarchical layout information.}
    \label{fig:intro}
    \vspace{-3mm}
\end{figure}

A geometric tree is a hierarchically arranged, tree-structured graph with nodes and edges that are spatially constrained, influencing their connectivity patterns. It is a particularly important data structure that is ubiquitous in different domains such as river geomorphology~\cite{brierley2013geomorphology, whipple2004bedrock}, neuron morphology~\cite{ascoli2007neuromorpho,parekh2013neuronal}, and vascular vessels~\cite{eguiluz2005scale}. Geometric trees are highly complex, non-linear structures and thus cannot be processed directly by common math and statistical tools. This makes representation learning on geometric trees a fundamental necessity in order to further apply them to downstream tasks such as classification, clustering, and generation. Although tree-structure representation learning has been extensively researched by techniques including sequence-based models, such as Tree-LSTM~\cite{tai2015improved} and graph neural networks~\cite{cao2020comprehensive,hamilton2017inductive,hamilton2017representation,kipf2016semi}, they are not able to jointly consider the geometric information that is coupled with the hierarchy and topology that are core properties of geometric trees. The importance of these attributes can be illustrated through real-life examples: as shown in Fig. \ref{fig:intro}(a), for a pyramidal neuronal cell, the closer to the cell body a branch is, the more curvature it exhibits. Similarly, as shown in Fig. \ref{fig:intro}(b), the node degree within a watershed's tree structure is indicative of the breadth of its corresponding subtree's expansion. On a theoretical level, as shown in Fig. \ref{fig:intro}(c), three geometric trees can be isomorphic if geometric information and hierarchy information about the levels from the root are not jointly considered.

Although recently some progress has been made with spatial graphs~\cite{schutt2017schnet,klicpera2020directional, zhang2021representation}, they cannot be used to directly handle geometric trees as they overlook the hierarchical ordering of nodes and edges, which, as mentioned above, is crucial. Therefore, this paper focuses on developing a method that can learn the representations of geometric trees by preserving their geometric, topological, and hierarchical ordering, as well as their interplay. To achieve this, three aspects need to be addressed: \textbf{1. How do we represent the node and its geometric and topological context in geometric trees?} To address this, our study introduces a novel representation learning model named \underline{G}eometric \underline{T}ree Branch \underline{M}essage \underline{P}assing (GTMP) that is specifically designed for geometric trees. At the core of our model is a unique message passing neural network that operates on the tree branches. The network is provably capable of preserving the entire geometric structure and ordering layout of trees, thereby ensuring an \textit{information-lossless} representation. Additionally, GTMP inherently supports crucial symmetry invariant properties, including rotation and translation invariance, enhancing its applicability to a wide range of geometric tree analyses. \textbf{2. How do we reflect the hierarchical patterns of geometric trees in the learned representations?} Existing methods tend to be either fully permutable or non-permutable; however, the hierarchical nature of geometric tree structures requires a partially permutable pattern. This pattern requires the preservation of parent-child ordering, whereas the ordering among siblings, for instance, does not need to be as rigidly preserved. To tackle this issue, we introduce a partial ordering constraint module that enforces a strict directional relationship in the embedding space for parent-child pairs to reflect their hierarchical order. \textbf{3. How do we learn the representations with label scarcity?} Unsupervised or self-supervised learning is often used nowadays for geometric data (e.g., graphs, images, spatial, etc.) representation learning. However, the inductive bias for self-supervised learning needs to be customized to specific geometric data types, and those pertaining to geometric trees  are significantly underexplored. To address this issue, a novel \underline{G}eometric \underline{T}ree \underline{S}elf-\underline{S}upervised \underline{L}earning (GT-SSL) framework is introduced through an innovative subtree-growing-guided objective, which aims to align the observed subtree and the expected subtree by its root.

%needs to decide what points you want to express first:
%[accuracy outperformed by xx\%?]
%[many metrics?]
%[many datasets?]
%[ diversity of experiment?]
%[what facts you demonstrated and how you demonstrated that?]
%...

To validate the effectiveness of our approach, comprehensive evaluations across eight real-world datasets have been conducted to demonstrate that our GTMP surpasses current leading geometric representation learning approaches by an average of over 15.6\% in AUC scores. Furthermore, our proposed self-supervised learning strategy, GT-SSL, further improves the prediction performance by an additional 13.9\% improvement on average. This underscores the effectiveness of our self-supervised learning approach in further refining the representations of geometric trees.

The remainder of this work is organized as follows: We begin by discussing existing relevant studies in the Section~\ref{sec:related_works}. This is followed by a formal problem definition in the Section~\ref{sec:preliminary}. Subsequently, in the Section~\ref{section:method}, the proposed GTMP framework is described. We conclude with comprehensive experimental results, assessing aspects such as effectiveness and efficiency in Section~\ref{sec:results}.

\section{Related Work}\label{sec:related_works}

\subsection{Deep Learning for Tree-Structured Data}
In recent years, there have been several methods proposed to more specifically handle the hierarchical nature of trees. For instance, the Tree-LSTM (Long Short-Term Memory) is an extension of the traditional LSTM architecture designed to handle hierarchical tree structures~\cite{tai2015improved}. Each node in the tree has its own memory cell and gate mechanisms that control the flow of information, enabling the model to retain context and information over longer hierarchical distances. On the other hand, tree transformers integrate tree structures with more typical transformer-based architectures, usually by restricting attention heads to focus on specific constituents of the tree~\cite{wang2019tree}. Tree-LSTM and tree transformers have been successfully applied to tasks involving tree-structured data, especially those in the NLP space~\cite{chen2016enhancing,shiv2019novel,harer2019tree}. However, these models only focus on the hierarchical nature of tree-structured data and are not explicitly designed for geometric trees that also include significant geometric and spatial data.

More generally, Graph Neural Networks (GNN) have also been adapted for use with tree-structured data. For instance, Qiao \textit{et al.} \cite{qiao2020tree} proposes T-GNN, a tree structure-aware graph neural network composed of two primary parts: an integrated hierarchical aggregation module, combining GNN with gated recurrent units, and a relational metric learning module to transform multi-hop neighborhood information. Hyperbolic GNNs (HGNN) leverage hyperbolic space to effectively model hierarchical and tree-like structures and capture long-range dependencies ~\cite{liu2019hyperbolic}.

\subsection{Spatial Networks}
Research into spatial networks has been ongoing for decades~\cite{barthelemy2011spatial}, dating back almost 50 years to Haggett and Chorley’s work developing some of the first models to characterize spatial networks~\cite{haggett1969network}. As improvements have been made with more complex networks~\cite{erdos1960evolution}, more modern quantitative advancements have emerged in fields such as transportation networks~\cite{banavar1999size}, urban mobility~\cite{chowell2003scaling}, biology~\cite{eguiluz2005scale}, spatial social dynamics~\cite{johnson2003spatial,wang2022invertible}, and computational chemistry~\cite{gilmer2017neural}.

As deep learning has progressed, numerous works have attempted to extend traditional deep learning methods to geometric data such as graphs~\cite{cao2020comprehensive,du2021graphgt,hamilton2017inductive,hamilton2017representation,kipf2016semi,ling2023deep,wang2022deep,zhang2021representation,zhang2023curriculum}. %For instance, the success of convolutional neural networks has inspired work on generalizing convolution and applying it to graphs~\cite{krizhevsky2012imagenet}. Some methods of graph convolution are spectral-based, such as ChebNet~\cite{tang2019chebnet}, while others are spatial-based, such as Graph Convolutional Networks (GCN) and Graph Isomorphism Networks (GIN)~\cite{niepert2016learning,xu2018powerful}, where convolution operations are primarily based on message-passing and pooling.
Many existing graph representation methods, however, tend to overlook hierarchies and the interplay between spatial and topological properties. Deep learning has also been applied to spatial data separately, such as PointNet for 3D point clouds or 3D Convolutional Neural Networks (CNN) for 3D voxel grids~\cite{qi2017pointnet,maturana2015voxnet,wu20153d}, but there is limited work conjoining graph and spatial representations, especially when applied to geometric trees where both hierarchical and spatial information are crucial.

\subsection{Self-Supervised Learning for Graph Data}
Self-supervised learning has emerged as a powerful paradigm for learning from graph-structured data, enabling models to leverage the rich, relational information inherent in graphs without requiring explicit labels. This approach has shown promise in extracting meaningful patterns and representations from graph data, pivotal across various domains such as social network analysis, chemical compound identification, and biological network interpretation~\cite{hamilton2020graph, velickovic2019deep}. Central to self-supervised learning on graphs is the design of pretext tasks that encourage the model to uncover and exploit the underlying structure and properties of the graph. Techniques such as predicting the presence of edges, estimating node properties based on local neighborhood structures, or reconstructing graph segments serve as effective learning signals~\cite{hu2020strategies, jing2020self}. 

However, existing self-supervised learning methods for graph-structured data often fall short in addressing the unique challenges of geometric tree self-supervised learning due to their limited focus on topology and neglect of geometric and hierarchical specificity. The unique hierarchical structures and the requirement of spatial invariance in geometric trees necessitate specialized approaches. %Consequently, there is a critical need for developing novel self-supervised learning frameworks that are tailored to capture the intricate relationships and spatial configurations inherent in geometric trees.
\section{Preliminaries and Problem Formulation}\label{sec:preliminary}
In this section, we first formalize geometric trees and the problem of representation learning on geometric trees, then we introduce the challenges with solving this problem.

Geometric trees (also known as spatial trees~\cite{barthelemy2011spatial}) are tree-structured graphs for which the nodes and edges of the tree occupy positions in a Euclidean space. The geometric constraints may significantly affect their tree topological patterns. Geometric trees are ubiquitous in the real world, such as river systems~\cite{whipple2004bedrock}, neurons~\cite{ascoli2007neuromorpho}, human blood vessels~\cite{eguiluz2005scale}, and mobility networks~\cite{chowell2003scaling}, where geometric and tree-structured network properties are intricately linked. For example, the structure of nerve endings is often jointly determined by both their upstream neurons and the local chemical environment~\cite{chung2015astrocytes}.

%(((((Geometric tree graphs (also known as spatial tree networks) are tree-structured graphs for which the nodes and edges of tree are embedded in a geometric space. Geometric tree graphs are ubiquitous in real-world scenarios, ranging from river systems and neurons to human blood vessels and mobility networks, where geometric and network properties are intricately intertwined. For example, the architecture of nerve endings is often collectively shaped by both their upstream neurons and the local chemical environment.)

A geometric tree can be formally represented as \(S=(T,P)\), where \(T=(V,E)\) symbolizes the tree-structured graph. In this representation, \(V\) is the set of \(N\) nodes, and \(E\), a subset of \(V \times V\) with \(|E|=|V|-1\), represents the \(N-1\) edges. Each edge \(e_{ij} \in E\) connects a \textit{parent} node \(v_i\) to a \textit{child} node \(v_j\) in \(V\), establishing a hierarchical relationship where \(v_i\) is the parent of \(v_j\).
In this structure, starting from any node \(v_i \in V\), it is impossible to traverse a \textit{path} (e.g., \(v_i \rightarrow v_{i1} \rightarrow v_{i2} \rightarrow v_i\)) that forms a loop, ensuring the acyclic nature of the tree.
A rooted tree, denoted as \(T_i\), originates from any node \(v_i\). If a node \(v_j\) is a \textit{descendant} of node \(v_i\), then its rooted tree \(T_j\) forms a \textit{subtree} within the larger rooted tree \(T_i\). In this hierarchical arrangement, node \(v_i\) is recognized as the \textit{ancestor} of node \(v_j\). 
The set $P$ represents the spatial coordinates, defined as $P=\{(x_i,y_i,z_i) | x_i, y_i, z_i \in \mathbb{R}\}$, within the Cartesian coordinate system. For each node $v_i \in V$, its spatial position is denoted by the coordinate tuple $(x_i, y_i, z_i) \in P$. 

%(((((Furthermore, to address natural and common requirements for invariance to rotation and translation in spatial data, as highlighted in X, we introduce a set of...)

The primary objective of this paper is to learn the representation $f(S)$ for geometric trees $S=(T,P)$, aiming to achieve a strong discriminative capability for unique geometric tree structures and to capture significant symmetry properties. This goal presents several unique challenges:

1) \textbf{It is difficult to jointly preserve both geometric and tree topological properties in geometric trees.} As shown in the example Figure~\ref{fig:intro}, the synergy between geometric attributes and tree topology is essential for generating distinguishable representations. %Furthermore, to meet the widespread need for rotation and translation invariance in geometric structure, as highlighted in~\cite{fuchs2020se,klicpera2020directional}, we introduce a set of \textit{rotation- and translation-invariant functions} for geometric trees. These functions are defined to maintain consistency under transformations, such that $f(T, \Psi(P)) = f(T, P)$ for all transformations $\Psi \in \mathrm{SE(3)}$. This group, $\mathrm{SE(3)}$ denotes the continuous Lie group responsible for rotation and translation transformations in three-dimensional space, $\mathbb{R}^3$.

%2) \textbf{Difficulty in incorporating the unique hierarchical relationships within geometric tree in the representations.} The hierarchical relationships within geometric trees dictate how individuals interact and influence each other, which is pivotal for understanding the overall structure and function of the tree. Traditional methods struggle to accurately represent these complex, layered relationships, necessitating advanced approaches that can encapsulate the hierarchical interactions (e.g. ancestor and descendent pair) within the tree structure, ensuring that the learned representations accurately mirror the intrinsic hierarchical organization.

2) \textbf{It is difficult to embed geometric hierarchy patterns.} Existing works either consider or ignore node permutation. However, geometric tree hierarchy requires capturing them in some patterns (e.g., parent-child) but not for others (e.g., siblings).

3) \textbf{It is difficult to label sufficient geometric tree data.} The scarcity of labeled data in specialized areas, such as neuron data, necessitates alternative approaches such as self-supervised learning. However, specialized self-supervised learning strategies for geometric trees are underexplored. %The scarcity of labeled data in real-world contexts severely limits the ability to train models effectively in a fully supervised learning manner. This issue is particularly pronounced in specialized domains such as neuron data, where obtaining comprehensive, accurately labeled datasets can be prohibitively expensive or logistically infeasible. Consequently, there is a pressing need for alternative training strategies that can overcome this hurdle, such as employing self-supervised learning or transfer learning methods. Nevertheless, crafting self-supervised learning objectives that not only leverage the unique properties of geometric trees but are also adaptable across various domains presents a complex challenge.
\section{Method}\label{section:method}
In order to develop a novel geometric tree representation learning method by addressing the challenges outlined above, we propose a new representation learning model named \underline{G}eometric \underline{T}ree Branch \underline{M}essage \underline{P}assing (GTMP) to fully exploit the interplay between geometric and tree-topological structures. In addition, a novel Geometric Tree Self-Supervised Learning (GT-SSL) framework is introduced to extract the customized geometric tree properties without any supervision labels. Specifically, to discriminate geometric trees, especially for the spatial tree joint patterns, we propose a new message passing scenario that aggregates the geometric information via tree branches, which is elaborated on in Section~\ref{subsection:GTMP}. This scenario preserves the geometric structure of tree information with theoretical guarantees on the invariance to $\mathrm{SE(3)}$-symmetric transformations and \textit{spatial-information-lossless}. To address the issue of insufficient labels, we developed two self-supervised learning objectives that are tailored for intrinsic geometric tree structures. Specifically, to incorporate the underlying hierarchical relationships, a partial ordering constraint over the parent-child pair embeddings is introduced in Section~\ref{subsection:parital_order}. This implies that a node's embedding should maintain a clear directional relationship with its subtree nodes to accurately represent the hierarchical structure. To introduce geometric tree-specific inductive bias as a self-supervised learning target, we further propose a top-down subtree growth learning process. As discussed in Section~\ref{subsection:generative_subtree}, our goal is to align the observed geometric structure of the subtree with the structure anticipated by its root. %The idea of choosing the generative learning process is attributed to the nature of growth process of a geometric tree typically starts from high level nodes to low level nodes. For example, in a river system, the geometric structure of tributary is typically determined by its rooted mainstream. 

%\ZZ{Think "partial" is spelled wrong in the subsection reference name; reply: it does not matter, but thanks} 

\subsection{Tree Branch Geometric-Topology Information Representation Learning} \label{subsection:GTMP}

\begin{figure}[t]
    \centering
    \includegraphics[width=1.\linewidth]{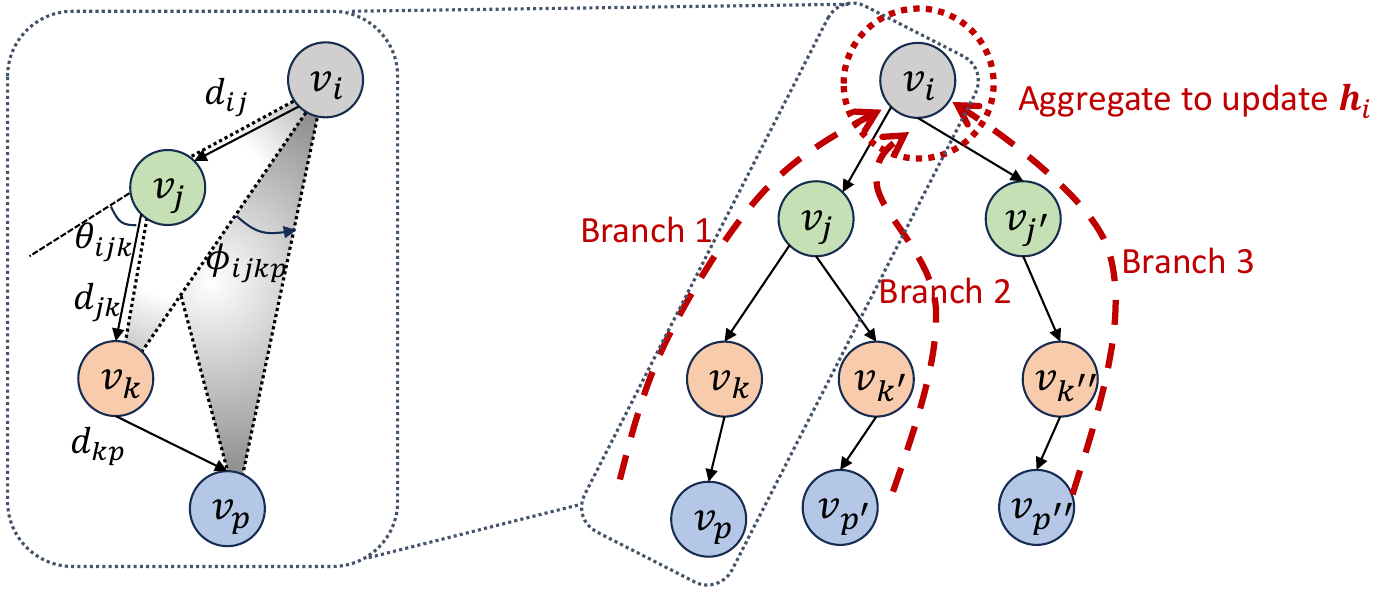}
    \vspace{-5mm}
    \caption{Illustration of the GTMP model. The geometric information is first extracted on each length-three branch starting from node $v_i$, then they are aggregated with other node information to update the node embedding $\mathbf{h}_i$.}
    \label{fig:GTMP}
    \vspace{-2mm}
\end{figure}

To effectively tackle the complexities of geometric tree representation learning, we have first developed an innovative message-passing technique known as Geometric Tree Branch Message Passing (GTMP). As illustrated in Figure~\ref{fig:GTMP}, this approach first harnesses the interplay between geometric properties and topological structures, enabling the computation of a comprehensive geometric-topology information representation for all branches originating from a tree node. Subsequently, we employ a neural network designed in a message-passing fashion, tailored specifically to account for the hierarchical ordering inherent in the tree branch structure. Our method systematically aggregates spatial information along an ordered tree branch. This strategy not only maintains the integrity of the tree's geometric structure but also assures robustness against $\mathrm{SE(3)}$-symmetric transformations, thereby enhancing the discriminative power of the process. 

Formally, the spatial information of a geometric tree with $N$ nodes can be expressed as a set of Cartesian coordinates $P=\{(x_i,y_i,z_i)|x_i,y_i,z_i \in \mathbb{R}\}_{i=1}^N$. It can also be represented as $\mathbf{P}\in\mathbb{R}^{N\times 3}$ in a matrix form. The set of all \textit{length $n$ tree paths} starting from node $v_i$ to its descendant nodes can be represented as $\pi_n^i$. In particular, a \textit{length three branch} $v_i \rightarrow v_j \rightarrow v_k \rightarrow v_p$ can be expressed as $\pi_{ijkp}\in \Pi_3^i$, where $v_i$ is the parent node of $v_j$, $v_j$ is the parent node of $v_k$, and $v_k$ is the parent node of $v_p$. Given a \textit{length three branch} $\pi_{ijkp}\in \Pi_3^i$, the proposed spatial information representation can be expressed as
\begin{equation}\label{eq:representation}
    (d_{ij},d_{jk},d_{jp}, \theta_{ijk}, \theta_{ijp}, \varphi_{ijkp}),
\end{equation}
where 
\begin{equation}    
\begin{split}
      d_{ij} & = ||\mathbf{P}_{ij}||_2, \quad d_{jk} = ||\mathbf{P}_{jk}||_2, \quad d_{jp} = ||\mathbf{P}_{jp}||_2, \\
      \theta_{ijk} & = \arccos\left(\left\langle \frac{\mathbf{P}_{ij}}{d_{ij}}, \frac{\mathbf{P}_{jk}}{d_{jk}} \right\rangle\right), \quad \theta_{ijp} = \arccos\left(\left\langle \frac{\mathbf{P}_{ij}}{d_{ij}}, \frac{\mathbf{P}_{jp}}{d_{jp}} \right\rangle\right), \\
      \varphi_{ijkp} & = \mathrm{\delta}(\mathbf{n_{ijk}}, \mathbf{n_{ijp}}, \mathbf{P}_{ij}) \cdot \arccos\left(\left\langle \mathbf{n_{ijk}}, \mathbf{n_{ijp}} \right\rangle\right), \\
      \mathbf{n_{ijk}} & = \frac{\mathbf{P}_{ij} \times \mathbf{P}_{jk}}{||\mathbf{P}_{ij} \times \mathbf{P}_{jk}||_2}, \quad \mathbf{n_{ijp}} = \frac{\mathbf{P}_{ij} \times \mathbf{P}_{jp}}{||\mathbf{P}_{ij} \times \mathbf{P}_{jp}||_2},
\end{split} 
\label{eq:representation_elaborate}
\end{equation}
where $\mathrm{\delta}(\mathbf{n_{ijk}}, \mathbf{n_{ijp}}, \mathbf{P}_{ij})= \left\langle \frac{ \mathbf{n_{ijk}} \times \mathbf{n_{ijp}} }{ ||\mathbf{n_{ijk}} \times \mathbf{n_{ijp}}||_2 }, \frac{ \mathbf{P}_{ij} }{ ||\mathbf{P}_{ij}||_2 } \right\rangle.$

\textbf{Theorem 1.} \textit{The defined distances $d_{ij} \in [0,\infty)$, angles $\theta_{ijk} \in [0, \pi)$, and torsions $\varphi_{ijkp} \in [-\pi, \pi)$ exhibit rigorous invariance under any rotation and translation transformations $\mathcal{T}\in \mathrm{SE(3)}$.}

The proof of Theorem 1 is straightforward and provided in Appendix~\ref{sec:proof}. Notably, the representation formulated in Equation~\ref{eq:representation} not only achieves invariance to rotation and translation, but also simultaneously preserves essential information for reconstructing the original geometric tree structure under weak conditions.

\textbf{Theorem 2.} \textit{Given a geometric tree $S=(T,P)$, where $T$ denotes a tree structure with a minimum depth of $\zeta \geq 3$, if the Cartesian coordinates for any set of three non-collinear, connected nodes $(v_j,v_k,v_p)$ within a \emph{length three branch} $\pi_{ijkp}$ starting from node $v_i$ are known, then the Cartesian coordinates $P$ of the entire tree can be accurately determined. This determination is based on the representation outlined in Equation~\ref{eq:representation}.}

The proof to this theorem is a consequence of the following theorem, which is proved in Appendix~\ref{sec:proof}.

\textbf{Theorem 3.} \textit{Given Cartesian coordinates of three non-collinear connected nodes $(v_j,v_k,v_p)$ in a \textit{length three branch} $\pi_{ijkp}$ of one node $v_i$, the Cartesian coordinate $P_i$ of node $v_i$ can be fully determined by the representation defined in Equation ~\ref{eq:representation}.}\\

\textit{Proof of Theorem 2.} As stated in Lemma 1, the Cartesian coordinate of node $v_i$ can be determined by its connected neighbors $v_j,v_k,v_p$ in the path of $\pi_{ijkp}$. Due to the property of strong connectivity of graph $G=(V,E)$, we can repeatedly solve the coordinate of a connected node to the set of nodes with known coordinates. Thus, starting from an arbitrary length three path, the Cartesian coordinates P of the whole spatial network can be determined. 
\hfill$\Box$ 

Upon extracting geometric features as outlined in Equation~\ref{eq:representation}, the next step involves creating a convolutional strategy. This strategy aims to merge tree-topology and spatial information derived from both the geometric representation of branches and their structural layouts into a comprehensive tree node representation. The essential challenge lies in preserving the model's ability to discriminate while also maintaining the integrity of the tree's structure and geometric details during the aggregation process.

To achieve this, we propose the following operation to update the hidden state embeddings $h_i^{\ell}$ by aggregating the message passing along all length three branches $\Pi_3^i$ originating from node $v_i$:

\begin{equation}    
\begin{split}
      h^{(\ell+1)}_i  & = \sigma^{(\ell)}\Big(\mathrm{AGG}\big(\big\{{m}^{(\ell)}(\pi_{ijkp})| \pi_{ijkp}\in \Pi_3^i \big\}\big)\Big),  \\
\end{split} 
\label{equation:arch1}
\end{equation}
where $\sigma^{(\ell)}$ is an arbitrary nonlinear transformation function (e.g. multilayer perceptron) and $\mathrm{AGG}$ denotes a set aggregation function. 

In our model, the representation of a branch $\pi_{ijkp}$ at layer $\ell$ integrates both topological and geometric information to produce a comprehensive message. The foundation of this approach lies in the aggregation of node features and the geometric configuration of the branch.

The integration of node features with geometric data is accomplished through a function $\phi^{(\ell)}$, which combines the aggregated node features $\bar{m}^{(\ell)}$ with a transformation of the geometric information $\psi^{(\ell)}(\hat{m}(\pi_{ijkp}))$. The final message for the branch, $m^{(\ell)}(\pi_{ijkp})$, is thus given by:
\begin{equation*}
      {m}^{(\ell)}(\pi_{ijkp})   = \phi^{(\ell)}\Big(\bar{m}^{(\ell)}(\pi_{ijkp}),\psi^{(\ell)}\big(\hat{m}(\pi_{ijkp})\big)\Big), 
\end{equation*}
\begin{equation}    
\begin{split}
      \bar{m}^{(\ell)}(\pi_{ijkp}) & = (h^{(\ell)}_i, \alpha_1 h^{(\ell)}_j, \alpha_2 h^{(\ell)}_k, \alpha_3 h^{(\ell)}_p), \\
      \hat{m}(\pi_{ijkp}) & = (d_{ij},d_{jk},d_{jp}, \theta_{ijk}, \theta_{ijp}, \varphi_{ijkp}),
\end{split} 
\label{equation:arch2}
\end{equation}
where $\phi^{(\ell)}$ and $\psi^{(\ell)}$ are two nonlinear functions to extract the complicated coupling relationship between geometric and tree topology information. 
The aggregated node features, $\bar{m}^{(\ell)}(\pi_{ijkp})$, are computed as a weighted combination of the features from node $v_i$ and its descendants in order: $v_j$, $v_k$, and $v_p$. 
Here $\alpha_1$, $\alpha_2$, and $\alpha_3$ adjust the influence of each ancestor's features at layer $\ell$.

\paragraph{Time Complexity}
It is worth highlighting that while our specially designed branch message passing network architecture incorporates the aggregation of higher-order neighborhood information, it diverges significantly from other higher-order message passing neural networks in terms of computational complexity. Typically, these other networks exhibit a $O(|N|^Q)$ time complexity, where $Q$ represents the neighborhood order. In contrast, our message passing mechanism achieves a more efficient performance by maintaining linear time complexity $O(|N|)$ relative to the number of nodes. This key distinction underscores the efficiency and scalability of our approach, making it uniquely suited for scaling to large-size data without the exponential increase in computational demand often associated with higher-order processing.

\subsection{Hierarchical Relationship Modeling through Partial Ordering Objective Function}\label{subsection:parital_order}

To accurately represent the inherent hierarchical relationships among nodes in tree structures, we introduce a partial ordering constraint module. The fundamental concept behind this function is to constrain embeddings in such a way that the embedding of a node in the tree not only represents itself but also maintains a structured relationship over its subtree nodes in the embedding space. %This configuration is crucial to maintain the tree's intrinsic order and inter-node dependencies, aspects that are often overlooked in traditional graph embedding strategies. The preservation of this natural hierarchy ensures that the embedding reflects the structured and hierarchical complexity of the tree, thus promoting a more powerful and accurate representation.
%This method reflects the actual hierarchical structure of the tree diagram, similar to the trunk of a biological tree as the superior node, and its branches and leaves as subordinate nodes. Therefore, the embedding of the trunk presents a superior relational posture compared to the embedding of branches and leaves. 

\begin{figure}[t]
    \centering
    \includegraphics[width=1.\linewidth]{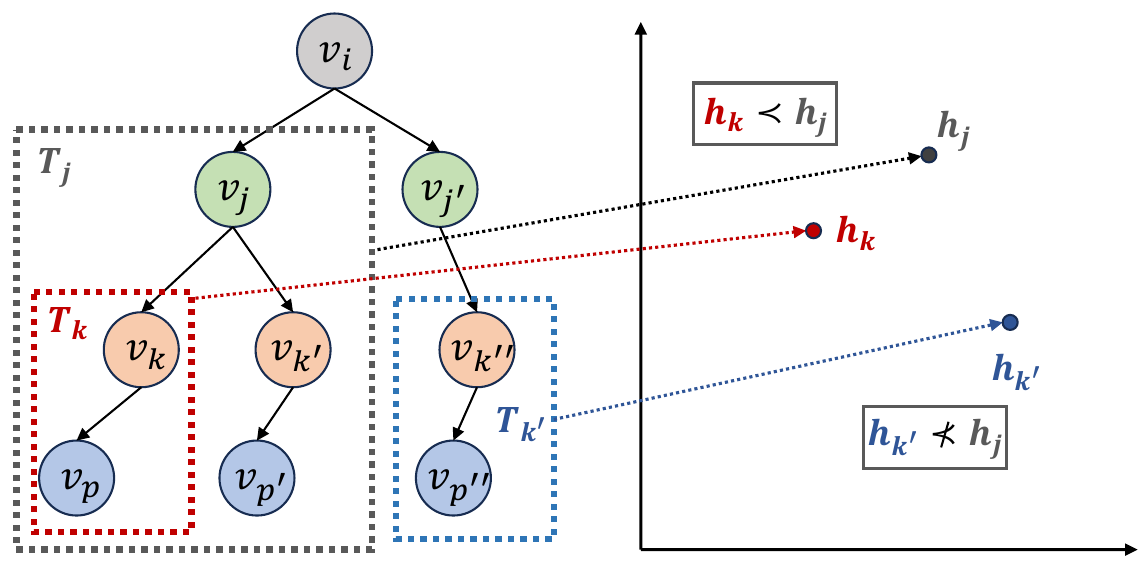}
    \vspace{-8mm}
    \caption{Illustration of the hierarchical relationship between tree nodes through partial ordering. In this scenario, $T_k$ is a subtree of $T_j$, establishing a partial ordering relationship between their respective subtree embeddings. Conversely, since $T_{k^{\prime}}$ does not constitute a subtree of $T_j$, there is no requirement for a partial ordering relationship between the embeddings of these two subtrees.}
    \label{fig:partial}
    \vspace{-5mm}
\end{figure}

To formally define the ordering constraint between node embeddings, we introduce the concept of partial ordering in the embedding space, ensuring that hierarchical relationships are accurately represented: as shown in Fig.~\ref{fig:partial}, if $T_j$ is a subtree of $T_i$, then the embedding $h_j$ of node $j$ has to be within the "lower-left" region of node $i$' embedding $h_i$:
\begin{equation}
    h_j[b] \leq h_i[b], \forall_{b=1}^{D} \quad \mathrm{iff} \quad T_j \subseteq T_i,
\end{equation}
where $D$ is the dimension of hidden embeddings and $[b]$ denotes the $b$-th dimension of hidden embeddings. 

To operationalize the above constraint into a function that can be optimized, we accordingly define the objective function for generating embeddings to utilize the max margin loss:
\begin{equation}
\mathcal{L}_{order} = \sum_{(h_i,h_j) \in \mathcal{P}} \max(0,  h_j - h_i) + \sum_{(h_i,h_j) \in \mathcal{N}} \max(0, \delta - \|h_i - h_j\|^2 ),
\end{equation}
where $\mathcal{P}$ and $\mathcal{N}$ denote the set of positive pairs and negative pairs in the minibatch where tree $T_j$ is a subtree of tree $T_i$. The term $\delta$ represents a margin that enforces a minimum distance between the embeddings of negative pairs compared to positive pairs, ensuring that $h_j$ (the embedding of the lower hierarchical node) is within the lower-left space to $h_i$ (the embedding of the higher hierarchical node), and is at least a margin distance $\delta$ apart for negative pairs.%\ZZ{Not sure what this means}

\subsection{Self-Supervised Learning via Subtree Growth Learning }\label{subsection:generative_subtree}
To tackle the challenge of insufficient training labels in real-world scenarios, we introduce an innovative Geometric Tree Self-Supervised Learning (GT-SSL) framework. While there are existing self-supervised learning objectives aimed at general graph data, they are insufficient when applied to geometric trees. The primary limitation is their inability to incorporate both intrinsic hierarchical relationships and coupled geometric-tree topology information into the self-supervised learning objectives. Our GT-SSL framework, by contrast, is specifically tailored to address these complexities, ensuring that the resulting representations fully reflect the unique structural and spatial characteristics of geometric trees. 

Our approach focuses on growing the geometric tree's information from the root node. We introduce a unique subtree-growth learning goal, which generates the entire geometric structure from top to bottom. This process unfolds iteratively, with each step predicting the geometric configuration of a node’s subtree based on the geometric structure of its ancestors. This approach is inspired by the observed natural growth patterns in geometric trees, where evolution occurs in a hierarchical manner, cascading from higher-level nodes down to lower-level ones. An illustrative example can be found in river systems, where the configuration of a tributary is largely influenced by the main river's structure and the characteristics of the surrounding geometric environment. %This generative method aims to replicate such growth dynamics, ensuring that the learned representations accurately reflect the fundamental developmental principles and environmental interactions characteristic of geometric trees.

To formalize our approach, for a node $v_i$ within a geometric tree $T$, we denote $\mathcal{C}(v_i)$ as the set of its child nodes and $\mathcal{A}(v_i)$ as the set of its ancestor nodes within the tree's hierarchy. The objective of our subtree growing process is to accurately predict the geometric structure of the child nodes, $\tilde{\mathcal{G}}(\mathcal{C}(v_i))$, given both the geometric structures of the node itself and that of its ancestors. This can be mathematically expressed as:
\begin{equation}
\tilde{\mathcal{G}}(\mathcal{C}(v_i)) = g\left(\mathcal{G}\left({v_i} \cup \mathcal{A}(v_i)\right)\right),
\end{equation}
where $g$ represents a learnable function designed to synthesize the geometric details of the child nodes based on the aggregated geometric information of $v_i$ and its ancestors. In our study, we prioritize predicting essential geometric features such as the distances between a node and its child nodes as well as the angles between the parent node, the current node, and its child nodes. 

Unfortunately, to feasibly represent the geometric structure information as $\mathcal{G}$, a significant challenge arises from the variable number of child nodes associated with any given node in a tree-structured graph, complicating the prediction of these geometric features. Simple aggregated indicators, such as averaging the distances, could be employed, but they risk obscuring the comprehensive geometric structure of the child nodes.

To overcome this limitation, we introduce an approach that involves converting the geometric features into the frequency domain. This transformation enables us to focus on predicting the frequency distribution of the geometric features across the child nodes, rather than attempting to directly predict specific values of the geometric features, thus avoiding the issue posed by the uncertain number of child nodes.

Formally, to represent the geometric information in a form that is amenable to pattern recognition and prediction, we expand these geometric features into the frequency space with radial basis functions. Mathematically, for a given node $v_i$ with child nodes $\mathcal{C}(v_i) = \{v_{i1}, v_{i2}, \dots, v_{in}\}$, the distances $\{d_{i1}, d_{i2}, \dots, d_{in}\}$ are expanded as follows:
\begin{equation}
    e_k (v_i) = \sum_{v_{ij}\in \mathcal{C}(v_i)} \exp (-\gamma \|d_{ij} - \mu_k\|^2),
\end{equation}
where $e_k$ denotes the $k$-th radial basis and $\mu_k$ is the corresponding distances. To obtain the distribution of geometric features over the radial basis, the ground truth distribution can be denoted as:

\begin{equation}
{\mathcal{G}}(\mathcal{C}(v_i)) = \left[ \sum_{v_{ij}\in \mathcal{C}(v_i)} e_k(v_i) \right]_{k=1}^{K} ,
\end{equation}
where $K$ is the total number of radial basis functions employed. Thus, the estimated distribution can be written as:

\begin{equation}
\hat{\mathcal{G}}(\mathcal{C}(v_i)) = g \left( \left[ \sum_{v_{ij}\in \mathcal{A}(v_i) \cup {v_i} } e_k(v_i) \right]_{k=1}^{K}\right).
\end{equation}

Subsequently, we formulate the objective function using Earth Mover's Distance (EMD) to measure the discrepancy between the estimated distribution $\tilde{\mathcal{G}}(\mathcal{C}(v_i))$ and the ground truth distribution $\mathcal{G}(\mathcal{C}(v_i))$:

\begin{equation}
\mathcal{L}_{generative} = \sum_{v_i\in V}\mathrm{EMD}\left(\tilde{\mathcal{G}}(\mathcal{C}(v_i)), \mathcal{G}(\mathcal{C}(v_i))\right),
\end{equation}
where EMD denotes the Earth Mover's Distance to quantify the cost of transforming the estimated distribution into the ground truth distribution.

Finally, the overall self-supervised learning objective function can be written as the combination of the subtree generative objective and the partial ordering function:

\begin{equation}
\mathcal{L}_{\mathrm{GT-SSL}} = \mathcal{L}_{generative} + \mathcal{L}_{order}
\end{equation}
\section{Experiments}\label{sec:results}
In this section, the experimental settings are introduced first in Section~\ref{exp:settings}, then the performance of the proposed method on the eight real-world datasets are presented in Section~\ref{exp:effective}. We further present the robustness test on our CL method against topological structure noise in Section~\ref{exp:robust}. We verify the transfer ability of framework through different datasets in Section~\ref{sec:exp-transfer}. The effectiveness of proposed components are measured through ablation studies in Section~\ref{exp:ablation}. In addition, we measure the parameter sensitivity in Section~\ref{exp:parameter} and running time analysis in Section~\ref{sec:time_analysis}. All experiments are conducted on a 64-bit machine with four NVIDIA A4000 GPUs (16 GB GDDR5). The proposed method is implemented with PyTorch~\cite{paszke2019pytorch} and the PyTorch-Geometric~\cite{Fey/Lenssen/2019} deep learning framework. The code to our work can be found in the Github repository: \url{https://github.com/rollingstonezz/KDD24_geometric_trees}.

\subsection{Experimental Settings}\label{exp:settings}

\begin{table*}[t]
\begin{adjustbox}{width=1.00\textwidth,center}
\begin{tabular}{l|l|ccccc|ccc}
\toprule
           &           & \multicolumn{5}{c|}{Neuron - Classification $(\uparrow)$}                                                                  & \multicolumn{3}{c}{River - Regression $(\downarrow)$}                                   \\ \midrule
           &           & lps-glia         & lps-inter        & lps-pc           & 5xfad-glia       & 5xfad-pc         & $\mu$          & $D$           & $r$            \\ \midrule
GCN        & Supervised  & 0.6063 $\pm$ 0.0235 & 0.5000 $\pm$ 0.0000 & 0.5000 $\pm$ 0.0000 & 0.7934 $\pm$ 0.0218 & 0.7543 $\pm$ 0.0611 & 0.1134 $\pm$ 0.0244 & 172.1041 $\pm$ 2.8023 & 87.6032 $\pm$ 1.2830 \\
           & GT-SSL & 0.9798 $\pm$ 0.0197 & 0.9568 $\pm$ 0.0134 & 0.9546 $\pm$ 0.0122 & 0.8131 $\pm$ 0.0111 & 0.8938 $\pm$ 0.0134 & 0.0941 $\pm$ 0.0321 & 95.5785 $\pm$ 2.3104  & 76.4080 $\pm$ 1.5892 \\
           & diff ($+/-$)   &     37.35\% & 	45.68\% &	45.46\% &	1.97\%	& 13.95\% &	17.01\% &	44.46\% &	12.77\%       \\ \midrule
GIN        & Supervised  & 0.6060 $\pm$ 0.0326 & 0.5173 $\pm$ 0.0388 & 0.5351 $\pm$ 0.0355 & 0.8204 $\pm$ 0.0083 & 0.7903 $\pm$ 0.0407 & 0.1872 $\pm$ 0.0150 & 172.5895 $\pm$ 3.6765 & 87.4989 $\pm$ 1.0998 \\
           & GT-SSL & 0.9784 $\pm$ 0.0147 & 0.9181 $\pm$ 0.0237 & 0.7943 $\pm$ 0.1235 & 0.8046 $\pm$ 0.0258 & 0.9343 $\pm$ 0.0109 & 0.1135 $\pm$ 0.0819 & 102.9059 $\pm$ 2.5516 & 73.3257 $\pm$ 1.3291 \\
           & diff ($+/-$)   &      37.24\% & 	40.08\% & 		25.92\%	 & 	-1.58\%	 & 	14.4\% & 		39.37\% & 		40.38\%	 & 	16.20\%    \\ \midrule
GAT        & Supervised  & 0.5878 $\pm$ 0.0411 & 0.5221 $\pm$ 0.0231 & 0.5000 $\pm$ 0.0000 & 0.8101 $\pm$ 0.0232 & 0.6576 $\pm$ 0.0981 & 0.2335 $\pm$ 0.0731 & 171.1645 $\pm$ 2.2507 & 89.0507 $\pm$ 2.5466 \\
           & GT-SSL & 0.5032 $\pm$ 0.0048 & 0.5000 $\pm$ 0.0000 & 0.5000 $\pm$ 0.0000 & 0.8318 $\pm$ 0.0177 & 0.8726 $\pm$ 0.0390 & 0.2752 $\pm$ 0.1201 & 141.8387 $\pm$ 5.1076 & 76.5700 $\pm$ 1.9842 \\
           & diff ($+/-$)   &       -8.46\%	& -2.21\%	& 0\%	& 2.17\%& 	21.5\%	& -17.86\%	& 17.13\%	& 14.02\%          \\ \midrule
PointNet   & Supervised  & 0.6308 $\pm$ 0.0835 & 0.7295 $\pm$ 0.0310 & 0.5508 $\pm$ 0.0431 & 0.8108 $\pm$ 0.0078 & 0.8960 $\pm$ 0.0140 & 0.1244 $\pm$ 0.0102 & 169.0530 $\pm$ 2.0981 & 85.9079 $\pm$ 2.4306 \\
           & GT-SSL & 0.9733 $\pm$ 0.0102 & 0.9543 $\pm$ 0.0140 & 0.6432 $\pm$ 0.0741 & 0.7707 $\pm$ 0.0316 & 0.9771 $\pm$ 0.0111 & 0.0754 $\pm$ 0.0107 & 94.3706 $\pm$ 2.7810  & 71.4937 $\pm$ 1.8721 \\
           & diff ($+/-$)   &    34.25\% &	22.48\% &	9.24\%	&-4.01\% &	8.11\% &	39.39\% &	44.18\%	& 16.78\%  \\ \midrule
SpatialNet & Supervised  & 0.7300 $\pm$ 0.0432 & 0.7445 $\pm$ 0.0419 & 0.5354 $\pm$ 0.0796 & 0.8614 $\pm$ 0.0113 & 0.8243 $\pm$ 0.0353 & 0.2335 $\pm$ 0.0418 & 151.0569 $\pm$ 2.5466 & 86.2461 $\pm$ 1.2461 \\
           & GT-SSL & 0.5010 $\pm$ 0.0025 & 0.5003 $\pm$ 0.0007 & 0.9232 $\pm$ 0.0460 & 0.8972 $\pm$ 0.0150 & 0.9764 $\pm$ 0.0164 & 0.0353 $\pm$ 0.0381 & 91.6590 $\pm$ 3.2311  & 66.9319 $\pm$ 3.1098 \\
           & diff ($+/-$)   &         -22.9\%	 &	-24.42\% &		38.78\%	 &	3.58\%	 &	15.21\%	 &	84.88\%	 &	39.32\% &		22.39\%          \\ \midrule
SchNet     & Supervised  & 0.7561 $\pm$ 0.0319 & 0.7597 $\pm$ 0.0286 & 0.6178 $\pm$ 0.0509 & 0.8994 $\pm$ 0.0105 & 0.9502 $\pm$ 0.0162 & 0.0342 $\pm$ 0.0072 & 154.4101 $\pm$ 3.2536 & 82.6100 $\pm$ 1.2830 \\
           & GT-SSL & 0.9698 $\pm$ 0.0120 & 0.9393 $\pm$ 0.0231 & 0.9612 $\pm$ 0.0202 & 0.9046 $\pm$ 0.0092 & 0.9893 $\pm$ 0.0083 & 0.0112 $\pm$ 0.0050 & 85.0109 $\pm$ 2.3832  & 37.8552 $\pm$ 1.4879 \\
           & diff ($+/-$)   &         21.37\%		 &	17.96\%		 &	34.34\%		 &	0.52\%		 &	3.91\%		 &	67.25\%		 &	44.94\%		 &	54.18\%   \\ \midrule
DimeNet    & Supervised  & 0.7049 $\pm$ 0.0620 & 0.8338 $\pm$ 0.0226 & 0.5601 $\pm$ 0.1020 & 0.9123 $\pm$ 0.0089 & 0.9544 $\pm$ 0.0120 & 0.0196 $\pm$ 0.0053 & 134.4048 $\pm$ 2.4952 & 80.9102 $\pm$ 1.2827 \\
           & GT-SSL & 0.9351 $\pm$ 0.0194 & 0.9627 $\pm$ 0.0176 & 0.9345 $\pm$ 0.0131 & \textbf{0.9540 $\pm$ 0.0059} & 0.9902 $\pm$ 0.0015 & 0.0077 $\pm$ 0.0033 & 80.2417 $\pm$ 2.0114  & 35.3140 $\pm$ 0.9910 \\
           & diff ($+/-$)   &     23.02\%	 &		12.89\%	 &		37.44\%		 &	4.17\%		 &	3.58\%		 &	60.71\%		 &	40.30\%		 &	56.35\%    \\ \midrule
SGMP       & Supervised  & 0.7599 $\pm$ 0.0442 & 0.8078 $\pm$ 0.0382 & 0.6231 $\pm$ 0.0512 & 0.8917 $\pm$ 0.0123 & 0.9839 $\pm$ 0.0034 & 0.0087 $\pm$ 0.0031 & \textbf{123.3789 $\pm$ 3.8385} & 68.0974 $\pm$ 2.1300 \\
           & GT-SSL & 0.9568 $\pm$ 0.0170 & 0.9709 $\pm$ 0.0142 & \textbf{0.9952 $\pm$ 0.0009} & 0.9333 $\pm$ 0.0085 & 0.9814 $\pm$ 0.0072 & \textbf{0.0033 $\pm$ 0.0012} & 76.7912 $\pm$ 1.4728  & 36.0287 $\pm$ 1.3890 \\
           & diff ($+/-$)   &     19.69\%	 &		16.31\% &			37.21\% &			4.16\%	 &		-0.25\%	 &		62.07\%	 &		37.76\%	 &		47.09\%   \\ \midrule
GTMP      & Supervised  & \textbf{0.7996  $\pm$ 0.0392} & \textbf{0.8529 $\pm$ 0.0370} & \textbf{0.6560 $\pm$ 0.0621} & \textbf{0.9417 $\pm$ 0.0123} & \textbf{0.9887 $\pm$ 0.0063} & \textbf{0.0052 $\pm$ 0.0024} & 125.4951 $\pm$ 2.8295 & \textbf{61.2353 $\pm$ 1.3177} \\
           & GT-SSL & \textbf{0.9836 $\pm$ 0.0096} & \textbf{ 0.9996 $\pm$ 0.0106} & 0.9872 $\pm$ 0.0013 & 0.9011 $\pm$ 0.0192 & \textbf{0.9992 $\pm$ 0.0004} & 0.0041 $\pm$ 0.0019 & \textbf{76.7699 $\pm$ 1.8740}  & \textbf{33.0287 $\pm$ 0.7680} \\
           & diff ($+/-$)   &      18.4\% &		11.87\%	 &	33.12\%	 &	-4.06\%	 &	1.05\%	 &	21.15\% &		38.83\% &		46.06\%       \\ \bottomrule
\end{tabular}
\end{adjustbox}
\caption{The main experimental results on neuron morphology and river flow network datasets. Here we present the performance of our GTMP method alongside other comparative methods within both supervised learning and our GT-SSL training approach. For each dataset, we highlight in bold both the best performance in the supervised learning context across all methods, and also the top performer within our GT-SSL training framework. Additionally, we show the percentage improvement in performance of all methods when leveraging our GT-SSL over traditional supervised learning settings. Specifically for the river flow network dataset, we use $\mu$ to represent the clustering coefficient, $D$ for spatial diameter, and $r$ for spatial radius.}
\label{table:main_results}
    \vspace{-7mm}
\end{table*}
\noindent\textbf{Datasets.}
To evaluate the performance of our proposed GTMP and comparison methods in real-world scenarios, eight geometric tree datasets are used in our experiments, which includes five neuron morphology datasets and three river flow network datasets across multiple tasks.

1) Neuron morphology: We conducted classification experiments using neuronal morphology data from NeuroMorpho.org, the largest online collection of 3D neural cell reconstructions contributed by hundreds of laboratories around the world~\cite{akram2018open}. Specifically, we constructed binary classification tasks between control cells and cells from two experimental conditions, 5xFAD and lipopolysaccharide injection (lps), where experimental condition was the target for prediction. All cells were mouse neural cells but tasks were split across three cell types: glia, interneurons (inter), and principal cells (pc). In total, we constructed five tasks: mouse 5xFAD glia versus mouse control glia; 5xFAD primary cells versus control primary cells; lps glia versus control glia; lps interneurons versus control interneurons; and lps primary cells versus control primary cells. A statistical description of the datasets is shown in Table~\ref{table:transfer_results}.

2) River flow networks: We also conducted regression experiments using publicly available river flow network data from the United States Geological Survey’s (USGS) National Hydrography Dataset (NHD)~\cite{USGSHydrography2019,USGSNHD2004}. We incorporated 2,231 river flow network tree samples with an average node count of 11,141 per tree. To facilitate prediction, we established three key geometric topology-coupled metrics as targets: the clustering coefficient, spatial diameter, and spatial radius.

\noindent\textbf{Comparison Methods.} 
 To the best of our knowledge, there has been little previous work directly handling geometric trees. Several advanced spatial graph networks have been developed to address generic spatial networks; among these, SchNet~\cite{schutt2017schnet}, DimeNet~\cite{klicpera2020directional}, and SGMP~\cite{zhang2021representation} are some of the closest related works to our approach and were selected as comparison methods. Additionally, we benchmarked our approach against three prominent graph neural network (GNN) methods— GCN~\cite{kipf2016semi}, GAT~\cite{velivckovic2017graph}, and GIN~\cite{xu2018powerful}— and two spatial neural network (SNN) techniques, PointNet~\cite{qi2017pointnet} and SpatialNet~\cite{danel2020spatial}. For GNN methods, Cartesian coordinates are provided as node attributes, while for SNN methods both node attribute and graph connectivity information are incorporated to ensure an equitable comparison. Further details on the benchmark models and comparison methodology are available in Appendix~\ref{apendix:experiment}.

\noindent\textbf{Implementation Details.} 
In the supervised learning configuration, all models utilize an identical architecture comprising three convolutional layers with the hidden dimension size set to 64. During the self-supervised representation learning pretraining phase, this architecture is maintained with three convolutional layers leading to final embeddings of 64 dimensions. For subsequent fine-tuning on specific tasks, a three-layer Multilayer Perceptron (MLP) is appended to the convolutional base, utilizing ReLU activation functions to enhance non-linear processing capabilities. To ensure a balanced evaluation across our proposed message passing mechanism and other GNN methods under comparison, we standardized the hyperparameter selection process. Detailed specifications of this process are delineated in Appendix~\ref{apendix:experiment}, providing transparency and facilitating reproducibility in our experimental setup.

We executed each experiment five times, subsequently averaging the results and computing the standard deviation. We adopted AUC score as the evaluation metric for the classification tasks on the neuron datasets due to the imbalanced distributions of all classes. On all runs in all tasks, the datasets were randomly divided into training, validation, and test sets with an 80:10:10 ratio, and identical hyperparameters were employed across all tasks for each dataset, except for the random seed that was responsible for the data split.

\vspace{-2mm}
\subsection{Effectiveness Results}~\label{exp:effective}
In this section, we first assess the performance of our GTMP method against competing approaches across the real-world datasets within a supervised learning manner. Additionally, we explore the efficacy of our GT-SSL framework to evaluate the quality of the generated representations in a pretrain-finetune manner over the same dataset. Given that the GT-SSL framework is designed to be a general approach applicable to various representation learning methods, we present the outcomes for both our GTMP approach and all other methods being compared. In this section, we pretrain and finetune the same datasets for implementing the GT-SSL framework, ensuring a fair comparison with the supervised learning results.

The comparison of AUC scores for the neuron morphology datasets and MAE results for the river datasets is provided in Table~\ref{table:main_results}. We summarize our observations on the effectiveness of the GTMP model and the GT-SSL training framework below:

\noindent (1) \textbf{Strength of GTMP model in learning effective geometric tree representations.} The supervised learning results demonstrate the strength of our proposed method, which consistently achieved the best results in seven out of eight datasets and securing the second-best result in the only dataset where the best performance was not attained. Specifically, our results outperformed the other comparison models by over $12.7\%$ on average for the neuron morphology datasets and $22.1\%$ on average for the river flow network datasets. The outcomes demonstrate that our GTMP model successfully leverages the specially designed branch message passing mechanism to generate representations, which significantly enhances performance on downstream supervised learning tasks.

\noindent (2) \textbf{Benefits of utilizing GT-SSL framework to enhance the quality of geometric tree representations.} Table~\ref{table:main_results} reveals that the GT-SSL's pretrain-finetune approach consistently enhances the performance of all representation learning models on 62 of the 72 total prediction tasks when compared to traditional supervised learning settings. Notably, the GT-SSL method surpasses supervised learning by an average margin of over $23.02\%$ across all tasks. This substantial improvement underscores the effectiveness of the GT-SSL pretraining framework in significantly enhancing the quality of learned representations via self-supervised learning objectives tailored to geometric tree structures. 

\noindent (3) \textbf{Integrating the GTMP model with the GT-SSL framework results in superior overall performance.} It is worth noting that the combination of our proposed GTMP model and GT-SSL framework shows a more competitive performance than any other combination of methods by achieving the best performance in six out of eight datasets. Specifically, our results outperformed the other comparison models by over $10.0\%$ on average for the neuron morphology datasets and $29.3\%$ on average for the river flow network datasets. The results demonstrate that integrating the GTMP model with the GT-SSL framework can successfully lead to state-of-the-art representation learning performance on geometric tree datasets.

\noindent (4) \textbf{Advantage of specialized spatial network representation learning methods over conventional graph and spatial neural networks.} It is also worth noting that the specialized spatial network representation learning methods (SchNet, DimeNet, SGMP, and GTMP) show a more competitive performance than both the vanilla graph neural network-based methods (GCN, GIN, and GAT) and the spatial network-based methods (PointNet and SpatialNet). Specifically, these specialized methods surpass traditional graph neural networks by an average of over $19.3\%$ in supervised learning contexts and $22.3\%$ when integrated with the GT-SSL framework; against spatial neural network approaches, they demonstrate an average improvement of $11.1\%$ in supervised settings and $23.7\%$ with GT-SSL. These results indicate that standard graph neural network and spatial neural network methods have limited capability to effectively discriminate patterns that require joint consideration of geometric information and tree topological information.

\begin{table}[t]
\begin{adjustbox}{width=1.00\columnwidth,center}
\begin{tabular}{cl|ccccc}
\hline
\multicolumn{1}{l}{}                         &     \multicolumn{1}{l}{}       & \multicolumn{5}{c}{Source}                                                                                                                                \\ \cline{3-7} 
\multicolumn{1}{l}{}                         &    \multicolumn{1}{l}{}        & \multicolumn{1}{l}{lps-glia} & \multicolumn{1}{l}{lps-inter} & \multicolumn{1}{l}{lps-pc} & \multicolumn{1}{l}{5xfad-glia} & \multicolumn{1}{l}{5xfad-pc} \\ 

                         &   \multicolumn{1}{c}{\# of Trees}         & {28,687} & {8,092} & {28,224} & {33,757} & {28,086} \\ 
           \multicolumn{2}{c}{Average \# of Nodes}              & {2,025} & {4,488} & {3,146} & {1,994} & {3,152} \\ \hline
\multicolumn{1}{c|}{\multirow{5}{*}{Target}} & lps-glia   & \underline{0.9836 }                      & 0.9983                        & 0.9987                     & 0.9231                         & \textbf{0.9999}                       \\
\multicolumn{1}{c|}{}                        & lps-inter  & 0.9111                       & \underline{0.9996}                        & 0.9936                     & 0.8627                         & \textbf{0.9999}                       \\
\multicolumn{1}{c|}{}                        & lps-pc     & \textbf{0.9995}                       & 0.9964                        & \underline{0.9872}                     & 0.9705                         & 0.9969                       \\
\multicolumn{1}{c|}{}                        & 5xfad-glia & 0.9357                       & \textbf{0.9981}                        & 0.9399                     & \underline{0.9011}                         & 0.9740                       \\
\multicolumn{1}{c|}{}                        & 5xfad-pc   & 0.9798                       & 0.9969                        & 0.9959                     & 0.8993                         & \textbf{\underline{0.9992}}                      \\ \hline
\end{tabular}
\end{adjustbox}
\caption{Transfer learning results. We underline the results where the source and target datasets are the same. Additionally, we highlight the best results for each target dataset.}
\label{table:transfer_results}
    \vspace{-8mm}
\end{table}

\subsection{Transfer Ability Analysis}\label{sec:exp-transfer}
We further investigate the transferability of our GT-SSL framework. In practical settings, the ability to deploy a model trained on one dataset to a new, unseen dataset without requiring retraining is highly beneficial. This strategy aims to address two main goals: (1) overcome the obstacle posed by insufficient data in the new dataset, which might hinder effective model training, and (2) save computational resources, as developing a model from the ground up demands significant time and resources. To assess how well our model adapts to new datasets, we initially pretrain the GTMP model using self-supervised learning objectives on source datasets. Subsequently, we finetune this pretrained model on target datasets to evaluate its performance in downstream task predictions.

The experimental results and sizes of datasets are shown in Table \ref{table:transfer_results}. (1) \textbf{Strong transfer ability across different source and target datasets.} It is evident that the transfer model, when applied from the source to the target dataset, can achieve performance on par with, or in some instances even surpassing, the model directly trained on the source dataset. Specifically, the discrepancy in average performance between scenarios where the source and target datasets are identical and those where they differ is a mere 0.91\%. (2) \textbf{Correlation between tree sizes and transfer performance.} More importantly, our findings reveal that pretraining on datasets with larger average tree sizes can significantly enhance transfer performance on target datasets. In particular, the average performance when pretraining on datasets characterized by relatively larger tree sizes (such as lps-inter, lps-pc, and 5xfad-pc) surpassed that of datasets with smaller tree sizes (such as lps-glia and 5xfad-glia) by an average of 5.51\%. Notably, the dataset with the largest average tree size, lps-inter, achieved an impressive average AUC score of 0.9979 across all datasets. These results underscore the ability of our proposed model to leverage larger dataset sizes for improving performance on unseen, relatively smaller-sized datasets. This capability presents a strategic advantage in addressing challenges related to data scarcity and computational constraints.

\begin{table}[t]
\begin{adjustbox}{width=1.\columnwidth,center}
\begin{tabular}{l|ccc|cc}
\hline
              & \multicolumn{3}{c|}{Neuron ($\uparrow$)}                         & \multicolumn{2}{c}{River ($\downarrow$)}             \\ \hline
              & lps-glia        & lps-inter       & 5xfad-pc        & Diameter         & Radius           \\ \hline
Supervised    & 0.7996          & 0.8529          & 0.9887          & 125.4951         & 61.2353          \\
GT-SSL        & \textbf{0.9836} & \textbf{0.9996} & \textbf{0.9992} & \textbf{76.7699} & \textbf{33.0287} \\
No Ordering   & 0.9769          & 0.9922          & 0.9981          & 78.8122          & 33.8345          \\
No Generative & 0.8235          & 0.8834          & 0.9847          & 108.5359         & 60.5332          \\ \hline
\end{tabular}
\end{adjustbox}
\caption{Ablation study results. NO Ordering refers to a variant that removes the partial ordering constraint module. NO Generative refers to another variant that removes the subtree growth learning module. The best result of each dataset is highlighted in bold. }
%\ZZ{Necessary to have all caps on "NO"? Yes, this is the most straightforward way}
\label{table:ablation}
    \vspace{-10mm}
\end{table}
\subsection{Ablation Studies}\label{exp:ablation}
This paper primarily concentrates on exploring the fundamental question of how effectively representation learning can leverage uniquely designed properties of geometric trees. Here, we investigate the impact of the proposed two self-supervised learning components of GT-SSL framework. We first consider a variant \textbf{No Ordering} that removes the \textit{Partial Ordering} module. To study the effectiveness of the proposed \textit{subtree growth learning} module, we further construct a variant \textbf{No Generative} that removes the corresponding module. Due to length constraints, we only present the results of five real-world datasets in Table~\ref{table:ablation}.

\noindent (1) Our full GT-SSL framework achieved the best performance on all five datasets. Specifically, the full model outperforms the variants \textbf{No Ordering} and \textbf{No Generative} by $10.3\%$ on average. In turn, these two variants exceeded the performance of the supervised learning model by an average of $13.3\%$. These outcomes confirm that incorporating \textit{partial ordering} and \textit{subtree growth learning} modules significantly enhances geometric tree representation learning tasks.

\noindent (2) The performance drops significantly when we remove the \textit{subtree growth learning} module, in comparison to removing the partial ordering module, which may indicate that this module plays a more critical role in understanding the joint geometric and tree topological properties towards learning powerful representations.

\begin{figure}[t]
    \centering
    \includegraphics[width=1.\columnwidth]{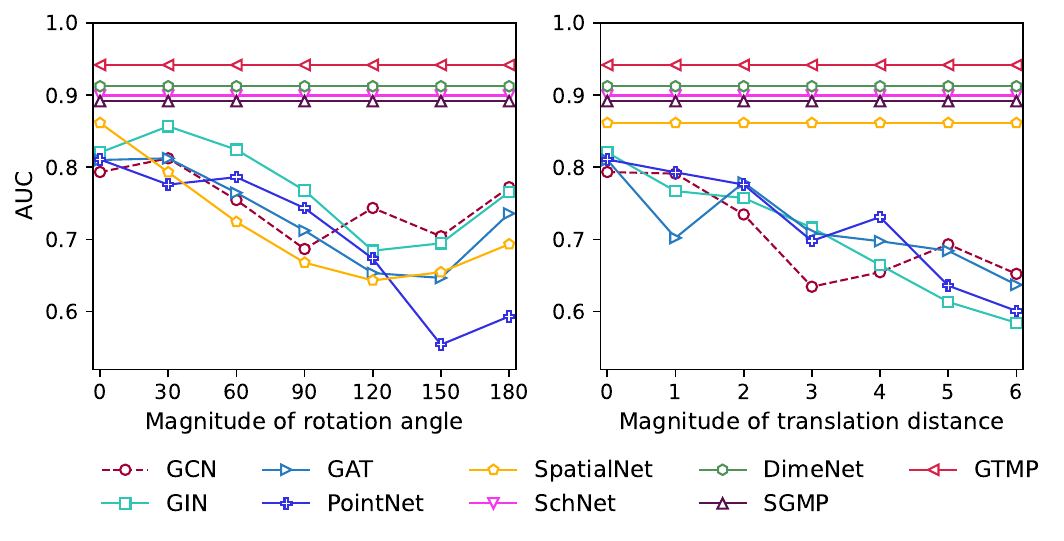}
    \vspace{-6mm}
    \caption{Robustness test on rotation and translation invariance by augmenting the data on the test set. The x-axis corresponds to the magnitude of the rotation angle (left) and translation distance (right) while y-axis shows the AUC scores. We can observe that our proposed GTMP model stays invariant to both translation and rotation transformations.}
    \label{fig:roto_translation}
    \vspace{-3mm}
\end{figure}

\begin{table*}[t]
\centering
\begin{adjustbox}{width=1.5\columnwidth,center}
\begin{tabular}{c|ccccc|ccc}
\hline
         & \multicolumn{5}{c|}{Neuron - Classification $(\uparrow)$} & \multicolumn{3}{c}{River - Regression $(\downarrow)$} \\ \hline
         & lps-glia  & lps-inter  & lps-pc  & 5xfad-glia  & 5xfad-pc & $\mu$           & $D$              & $r$              \\ \hline
$\{\alpha\}_1$ & 0.9836    & 0.9996     & 0.9872  & 0.9011      & 0.9992   & 0.0041          & 76.7699          & 33.0287          \\
$\{\alpha\}_2$ & 0.9779    & 0.9999     & 0.9764  & 0.8890      & 0.9985   & 0.0038          & 77.3823          & 32.0424          \\ \hline
lr=$1 \times 10^{-3}$  & 0.9836    & 0.9996     & 0.9804  & 0.8984      & 0.9991   & 0.0042          & 76.7699          & 33.0287          \\
lr=$5 \times 10^{-4}$  & 0.9810    & 0.9958     & 0.9872  & 0.9011      & 0.9992   & 0.0041          & 76.9832          & 33.8315          \\ \hline
bs=16  & 0.9836    & 0.9996     & 0.9804  & 0.9003      & 0.9991   & 0.0045          & 76.7699          & 33.0287          \\
bs=64  & 0.9849    & 0.9996     & 0.9751  & 0.9011      & 0.9991   & 0.0041          & 77.5353          & 34.1242          \\ \hline
\end{tabular}
\end{adjustbox}
\caption{Parameter sensitivity analysis. Here $\{\alpha\} = \{ \alpha_1, \alpha_2,  \alpha_3 \}$ denotes the hyperparameters to control the influence of each ancestor, lr is short for learning rate, and bs is short for batch size.
}
\vspace{-4mm}
\end{table*}

\begin{table*}[t]
    \centering
\begin{adjustbox}{width=1.5\columnwidth,center}
\begin{tabular}{l|cccccccccc}
\toprule
\# of Nodes    & 1,000  & 2,000  & 3,000 & 4,000  & 5,000  & 6,000  & 7,000   & 8,000   & 9,000   & 10,000  \\
\midrule
Execution Time & 1.905 & 3.209 & 4.850 & 6.272 & 8.102 & 9.826 & 11.371 & 13.248 & 15.462 & 17.084 \\
\bottomrule
\end{tabular}
\end{adjustbox}
    \caption{The efficiency analysis of the GTMP model, measured in seconds. }
    \label{tab:results}
\vspace{-4mm}
\end{table*}

\subsection{Parameter Sensitivity Analysis}\label{exp:parameter}
In this section we present the parameter sensitivity analysis for the GTMP model, focusing on the impact of three types of hyperparameters across eight datasets. Initially, we explore the effect of the hyperparameter set $\{\alpha\} = \{ \alpha_1, \alpha_2,  \alpha_3 \}$, which modulates the influence of each ancestor in Equation 4. The notation $\{\alpha\}_1$ signifies uniform weighting, achieved by setting each $\alpha$ value to 1, thereby distributing equal influence among ancestors. Conversely, $\{\alpha\}_2$ represents a decreasing weighting scheme, with $\alpha_1=1$ maintaining full influence, $\alpha_2 = 0.8$ indicating moderately reduced influence, and $\alpha_3= 0.5$ depicting significantly diminished influence.
Subsequently, we evaluate the model's sensitivity to the initial learning rate, comparing the effects of setting it to $1 \times 10^{-3}$ and $5 \times 10^{-4}$. Lastly, we assess the impact of selecting different batch sizes, as 16 and 64, on the model's performance.

\subsection{Rotation and Translation Invariant Test}
Similar to prior research~\cite{fuchs2020se, zhang2021representation}, our study evaluates the robustness against rotation and translation by uniformly applying these transformations to the input Cartesian coordinates. Due to length constraints, we present accuracy results for the classification task solely on the neuron dataset lps-inter, although the findings are consistent across all datasets. As depicted in Figure~\ref{fig:roto_translation}, our model demonstrates remarkable invariance to both translation and rotation, maintaining stable performance under these transformations. Comparable robustness is observed in models such as our proposed GTMP and comparison methods SchNet, DimeNet, and SGMP, attributable to their reliance on rotation and translation-invariant spatial features. In contrast, SpatialNet exhibits invariance to translation but not to rotation, as it employs only relative coordinates. This discrepancy underscores the critical advantage of incorporating rotation and translation-invariant features into models, as evidenced by the substantial performance decline of models lacking theoretical invariance when subjected to both rotation and translation transformations. This experiment highlights the significance of adopting models with built-in invariance to ensure consistent performance across varied spatial orientations and positions.

\subsection{Efficiency Analysis}\label{sec:time_analysis}
In this section, we further examine the efficiency of GTMP by measuring the relationship between the number of nodes in the trees, ranging from 1,000 to 10,000, and the execution time. The results are shown in Table \ref{tab:results}. The latter reflects the training time for one epoch of 1,000 trees, averaged over 100 runs. The Pearson correlation coefficient between the number of nodes and the execution time is 0.999, and the p-value is $1.76 \times 10^{-11}$. These results indicate a strong linear time complexity of our proposed method with respect to the number of nodes, demonstrating its efficiency and scalability.
\section{Conclusion}
%In this paper, we first propose Tree-Based Message Passing (TBMP) model to effectively learn coupled spatial-topology representation from geometric tree-structured data. Theoretical guarantees are given for the powerful representation ability on preseving sufficient geometric structure information. In order to address the insufficent labeled data issue and enable the transfer ability, we further propose the Self-Supervised Learning Framework for Geometric Trees (GT-SSL), which substantially enhances the performance of geometric tree representations through unique underlying hierarchies and tree-oriented geometric structures.  The integration of the TBMP model with the GT-SSL framework further accentuates its effectiveness, leading to state-of-the-art performance in various datasets.

In this paper, we introduce the Geometric Tree Message Passing (GTMP) model, designed to efficiently learn coupled spatial-topology representations from geometric tree-structured data. Theoretical guarantees are given to assure its ability to preserve essential geometric structure information. To overcome the challenge of insufficient labeled data and to enhance transferability, we also introduce the Self-Supervised Learning Framework for Geometric Trees (GT-SSL). This framework significantly improves geometric tree representations by leveraging their inherent hierarchies and tree-oriented geometric structures. The integration of the GTMP model with the GT-SSL framework further accentuates its effectiveness, leading to state-of-the-art performance on various datasets.

\section*{Acknowledgement}
This work was supported by the National Science Foundation (NSF) Grant No. 1755850, No. 1841520, No. 2007716, No. 2007976, No. 1942594, No. 1907805, a Jeffress Memorial Trust Award, Amazon Research Award, NVIDIA GPU Grant, and Design Knowledge Company (subcontract number: 10827.002.120.04). The authors acknowledge Emory Computer Science department for providing computational resources and technical support that have contributed to the experimental results reported within this paper.

\bibliography{sample-base}
\bibliographystyle{ACM-Reference-Format}

\appendix

\section{Mathematical Proofs}\label{sec:proof}
\textbf{Theorem 1.} \textit{The defined distances $d_{ij} \in [0,\infty)$, angles $\theta_{ijk} \in [0, \pi)$, and torsions $\varphi_{ijkp} \in [-\pi, \pi)$ exhibit rigorous invariance under any rotation and translation transformations $\mathcal{T}\in \mathrm{SE(3)}$.}

\noindent\textit{Proof of Theorem 1.} 
To demonstrate the invariance of distances, angles, and torsions with respect to translation transformations in Theorem 1, we note that these properties depend solely on relative coordinates, making them inherently immune to such transformations.

For rotation transformations characterized by $R$ in $SO(3)$, the 3D rotation group, we begin by establishing two fundamental identities:
\begin{align*}
 \langle R\mathbf{x}, R\mathbf{y} \rangle &= \langle \mathbf{x}, \mathbf{y} \rangle, \\
 (R\mathbf{x}) \times (R\mathbf{y}) &= R(\mathbf{x} \times \mathbf{y}).
\end{align*}
These identities allow us to confirm the invariance under rotations as follows:
\begin{align*}
 d_{ij} &= ||\mathbf{P}_{ij}||_2 = \sqrt{\langle \mathbf{P}_{ij}, \mathbf{P}_{ij} \rangle} = \sqrt{\langle R\mathbf{P}_{ij}, R\mathbf{P}_{ij} \rangle}, \\
 \theta_{ijk} &= \arccos\left(\frac{\langle \mathbf{P}_{ij}, \mathbf{P}_{jk} \rangle}{d_{ij}d_{jk}}\right) \\
 &= \arccos\left(\frac{\langle R\mathbf{P}_{ij}, R\mathbf{P}_{jk} \rangle}{d_{ij}d_{jk}}\right), \\
 \bar{\varphi}_{ijkp} &= \arccos\left(\frac{\langle \mathbf{n_{ijk}}, \mathbf{n_{jkp}} \rangle}{||\mathbf{n_{ijk}}||_2 ||\mathbf{n_{jkp}}||_2}\right) \\
 &= \arccos\left(\frac{\langle R\mathbf{n_{ijk}}, R\mathbf{n_{jkp}} \rangle}{||R\mathbf{n_{ijk}}||_2 ||R\mathbf{n_{jkp}}||_2}\right), \\
 \mathrm{\delta}(\mathbf{n_{ijk}}, \mathbf{n_{ijp}}, \mathbf{P}_{ij}) &= \frac{\langle \mathbf{n_{ijk}} \times \mathbf{n_{ijp}}, \mathbf{P}_{ij} \rangle}{||\mathbf{n_{ijk}} \times \mathbf{n_{ijp}}||_2 ||\mathbf{P}_{ij}||_2} \\
 &= \frac{\langle (R\mathbf{n_{ijk}}) \times (R\mathbf{n_{ijp}}), R\mathbf{P}_{ij} \rangle}{||(R\mathbf{n_{ijk}}) \times (R\mathbf{n_{ijp}})||_2 ||R\mathbf{P}_{ij}||_2}.
\end{align*}
This proof confirms that all considered elements retain their values under both rotation and translation, thus proving the invariance as stated in Theorem 1.
\hfill$\Box$ 

\textbf{Theorem 3.} \textit{Given Cartesian coordinates of three non-collinear connected nodes $(v_j,v_k,v_p)$ in a \textit{length three branch} $\pi_{ijkp}$ of one node $v_i$, the Cartesian coordinate $P_i$ of node $v_i$ can be fully determined by the representation defined in Equation ~\ref{eq:representation}.}\\

\noindent\textit{Proof of Theorem 3.}
Let $\mathbf{P}_j$, $\mathbf{P}_k$, and $\mathbf{P}_p$ denote the Cartesian coordinates of the nodes $v_j$, $v_k$, and $v_p$, respectively. Let $\mathbf{P}_i$ denote the Cartesian coordinate of node $v_i$, which we seek to determine.

Given the distance $d_{ij}$ between nodes $v_i$ and $v_j$; the angle $\theta_{ijk}$ between the vectors $\mathbf{P}_i - \mathbf{P}_j$ and $\mathbf{P}_k - \mathbf{P}_j$; the torsion angle $\bar{\varphi}_{ijkp}$ defined by the plane containing $\mathbf{P}_i$, $\mathbf{P}_j$, $\mathbf{P}_k$; and the plane containing $\mathbf{P}_j$, $\mathbf{P}_k$, $\mathbf{P}_p$:

1. The vector $\mathbf{v}_{ij} = \mathbf{P}_i - \mathbf{P}_j$ is such that $||\mathbf{v}_{ij}|| = d_{ij}$.

2. The angle $\theta_{ijk}$ satisfies the equation:
\[
\cos(\theta_{ijk}) = \frac{(\mathbf{P}_i - \mathbf{P}_j) \cdot (\mathbf{P}_k - \mathbf{P}_j)}{||\mathbf{P}_i - \mathbf{P}_j|| \, ||\mathbf{P}_k - \mathbf{P}_j||}.
\]

3. The torsion angle $\bar{\varphi}_{ijkp}$ involves the cross product of the normals to the planes $(\mathbf{P}_i, \mathbf{P}_j, \mathbf{P}_k)$ and $(\mathbf{P}_j, \mathbf{P}_k, \mathbf{P}_p)$:
\[
\sin(\bar{\varphi}_{ijkp}) = \frac{(\mathbf{n}_{ijk} \times \mathbf{n}_{jkp}) \cdot (\mathbf{P}_k - \mathbf{P}_j)}{||\mathbf{n}_{ijk} \times \mathbf{n}_{jkp}|| \, ||\mathbf{P}_k - \mathbf{P}_j||},
\]
where $\mathbf{n}_{ijk} = (\mathbf{P}_i - \mathbf{P}_j) \times (\mathbf{P}_k - \mathbf{P}_j)$ and $\mathbf{n}_{jkp} = (\mathbf{P}_k - \mathbf{P}_j) \times (\mathbf{P}_p - \mathbf{P}_j)$.

To determine $\mathbf{P}_i$, we reverse-engineer these relationships starting from known quantities $\mathbf{P}_j$, $\mathbf{P}_k$, and $\mathbf{P}_p$. Given $d_{ij}$, we find all possible $\mathbf{P}_i$ that satisfy the distance constraint. Among these, $\theta_{ijk}$ narrows the possibilities to a circle in the plane defined by $\mathbf{P}_i$, $\mathbf{P}_j$, and $\mathbf{P}_k$. Finally, $\bar{\varphi}_{ijkp}$ selects a unique $\mathbf{P}_i$ by specifying its orientation relative to $\mathbf{P}_p$.

Therefore, the coordinate $\mathbf{P}_i$ can be uniquely determined by solving these geometric constraints simultaneously, leveraging the non-collinearity of $\mathbf{P}_j$, $\mathbf{P}_k$, and $\mathbf{P}_p$ to ensure a unique solution exists.
\hfill$\Box$

\textbf{Theorem 2.} \textit{Given a geometric tree $S=(T,P)$, where $T$ denotes a tree structure with a minimum depth of $\zeta \geq 3$, if the Cartesian coordinates for any set of three non-collinear, connected nodes $(v_j,v_k,v_p)$ within a \emph{length three branch} $\pi_{ijkp}$ starting from node $v_i$ are known, then the Cartesian coordinates $P$ of the entire tree can be accurately determined. This determination is based on the representation outlined in Equation~\ref{eq:representation}.}

\noindent\textit{Proof of Theorem 2.}
As stated in Lemma 1, the Cartesian coordinate of node $v_i$ can be determined by its connected neighbors $v_j,v_k,v_p$ in the path of $\pi_{ijkp}$. Due to the property of strong connectivity of graph $G=(V,E)$, we can repeatedly solve the coordinate of a connected node to the set of nodes with known coordinates. Thus, starting from an arbitrary length three path, the Cartesian coordinates P of the whole spatial network can be determined. 
\hfill$\Box$

\section{Additional Experimental Settings}\label{apendix:experiment}

\subsection{Comparison Methods}
This section provides details on the comparison models used in our study.

\textbf{Graph Neural Networks (GNNs).}

\textbf{(i)~GIN.} Graph Isomorphism Networks (GIN), proposed by Xu et al.~\cite{xu2018powerful}, is a type of GNN known for its powerful capability to discriminate graph structures within the class of 1-order GNNs; 

\textbf{(ii)~GAT.} Graph Attention Networks (GAT)~\cite{velivckovic2017graph} leverages multi-head attention layers to facilitate information propagation across the graph;

\textbf{(iii)~GCN.} GCN~\cite{kipf2016semi} is a commonly used GNN model by using a localized first-order approximation of spectral graph convolutions;

\textbf{Spatial Neural Networks (SNNs).}

\textbf{(i)~PointNet.} PointNet~\cite{qi2017pointnet} processes pointwise features independently using multiple MLP layers and aggregates them into global features with a max-pooling layer;

\textbf{(ii)~SpatialNet.} SpatialNet~\cite{deng2018ppfnet} is a spatial deep learning framework designed to learn globally aware 3D descriptors;

\textbf{Spatial Networks.}

\textbf{(i)~SchNet.} SchNet~\cite{schutt2017schnet} is a model tailored for spatial networks predictions. It applies a continuous filter function to model distances between nodes and their nearest neighbors;

\textbf{(ii)~DimeNet.} DimeNet~\cite{klicpera2020directional} is another model focused on spatial networks. It incorporates directional information by aggregating messages from ``length two paths'', using a physically informed representation of distances and angles;

\textbf{(iii)~SGMP.} SGMP~\cite{zhang2021representation} is a deep learning approach tailored for spatial networks within Euclidean spaces. It enhances the model's interpretability and accuracy by incorporating directional information. This is achieved through the aggregation of messages from paths of length three, utilizing a physically-grounded representation that includes distances, angles, and torsions to encode spatial relationships.

\subsection{Searching Space for Hyper Parameters}
Number of epochs trained$:$ $\{50,100,200\};$\\
Size of each batch$:$ $\{16,32,64\};$\\
Initial learning rate for model$:$ $\{1e{-3}, 5e{-3}, 1e{-4}\};$\\
Number of convolution layers$:$ $\{3\};$\\
Dimension of hidden state$:$ $\{64\};$\\
Weight decay ratio$:$ $\{0.9\}$ (if loss on validation set does not decrease for 10 epochs).

\subsection{Example Visualization of Experimental Data}

\begin{figure}[h]
    \centering
    \includegraphics[width=0.9\linewidth]{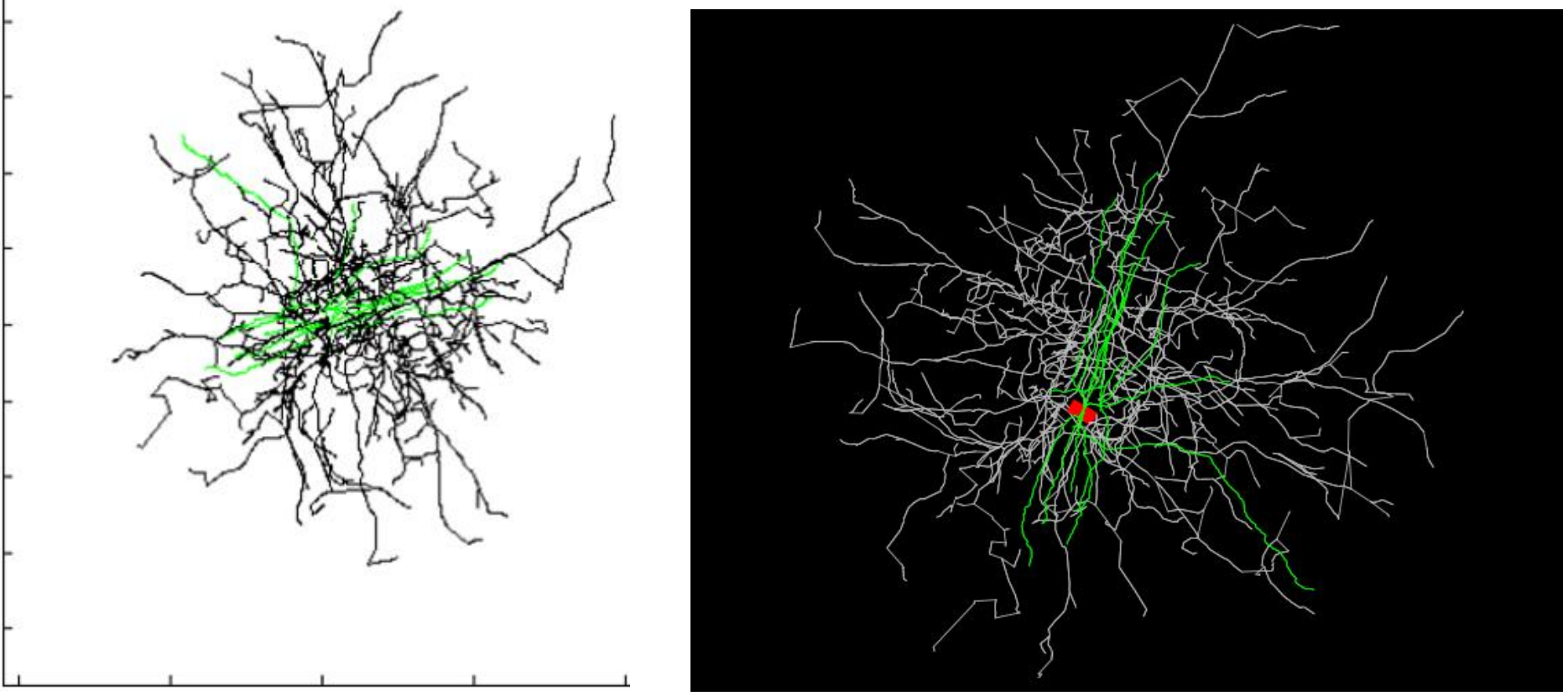}
    \caption{Images of example neural cell reconstruction from NeuroMorpho.org, which depict the reconstruction for the cell with NeuroMorpho.org ID NMO\_86952 ~\cite{ascoli2007neuromorpho,koch2016big}. The left image is a screenshot of the cell as viewed in the “Animation” feature on the cell’s corresponding NeuroMorpho.org page, and the right image is the cell’s representative image in the database. }
    \label{fig:enter-label}
\end{figure}

\end{document}